# DRiVE: Dynamic Recognition in VEhicles using snnTorch

Heerak Vora
*Dept. of Computer Engineering*
*Sardar Vallabhbhai Patel Institute of Technology, Vasad*
Anand, India
0009-0003-8277-8245

Param Pathak
*Dept. of Computer Engineering*
*Sardar Vallabhbhai Patel Institute of Technology, Vasad*
Anand, India
0009-0003-6419-0915

Parul Bakaraniya
*Dept. of Computer Engineering*
*Sardar Vallabhbhai Patel Institute of Technology, Vasad*
Anand, India
0000-0001-8072-6837

*Abstract* — Spiking Neural Networks (SNNs) mimic biological brain activity, processing data efficiently through an event-driven design, wherein the neurons activate only when inputs exceed specific thresholds. Their ability to track voltage changes over time via membrane potential dynamics, helps retain temporal information. This study combines SNNs with PyTorch's adaptable framework, snnTorch, to test their potential for image-based tasks. We introduce DRiVE, a vehicle detection model that uses spiking neuron dynamics to classify images, achieving 94.8% accuracy and a near-perfect 0.99 AUC score. These results highlight DRiVE's ability to distinguish vehicle classes effectively, challenging the notion that SNNs are limited to temporal data. As interest grows in energy-efficient neural models, DRiVE's success emphasizes the need to refine SNN optimization for visual tasks. This work encourages broader exploration of SNNs in scenarios where conventional networks struggle, particularly for real-world applications requiring both precision and efficiency.

## I. Introduction

Artificial intelligence (AI) advances in recent years have led to many developments among neural network architectures, each trying to mimic the complexity of biological neural systems. One such architecture Spiking Neural Networks (SNNs) stands out due to its ability to emulate temporal characteristics of biological neurons more closely than conventional and traditional artificial neural networks (ANNs). SNNs are based on encoding through spikes generated when a neuron's membrane potential surpasses a specific threshold. It is similar to the way in which neurons operate using electrical impulses known as action potentials [1]. This mechanism of firing allows SNNs to process information using timing of spikes – an approach known as temporal coding. SNNs can adapt the synaptic weights effectively using spike-timing-dependent plasticity, providing a robust framework for learning. To make working with and implementing SNNs easier, snnTorch proves to be a significant tool. It is a Python package designed to incorporate spiking neurons by extending the capabilities of PyTorch. [2]. This framework architecture supports exploration on both small and large networks through various neuron models with features for spike generation, data conversion and visualization. This integration helps conduct research and experiments requiring high performance in real time applications such as computer vision and robotics. There have been numerous attempts at providing effective models for image detection particularly utilizing convolutional neural networks (CNNs) and other traditional methods [3], [4]. These models often require substantial computational resources and energy consumption. SNNs offer an alternative in this aspect providing comparable performance while significantly reducing power consumption and enhancing computational efficiency [1]. This proves to be critical in applications such as image detection that may be further leveraged to construct real time systems and applications such as autonomous vehicles. Through this paper, we aim to utilize these capabilities of SNN using snnTorch implementation for vehicle detection. This represents an innovative approach wherein we explore whether employing SNNs can match the performance at a task that traditionally relies heavily on methods such as convolutional neural networks (CNNs). It helps us investigate if temporal processing can enhance the efficiency of present algorithms without compromising their performance. Such algorithms would be capable of operating under constrained resources which would be significant for applications such as autonomous vehicles.

## II. Literature review

SNN has very wide scale applications and is considered $3^{rd}$ generation of neural networks [5]. Research on the same has just been initialized and is still under progress. Very few researchers have studied this under-appreciated method and identified its core principles [6]. For facilitating seamless implementation of SNNs, there exist several frameworks which help in training and utilization, particularly snnTorch [2]. These frameworks may be used for many varied applications such as the usage of Lava framework for bayesian optimization [7], and resource allocation [8]. One such application of SNN, image detection, has few derived methodologies with considerable accuracy namely AMOS (80.97 %) [9], Spikformer V2 (94.80 %) [10], S-ResNet30 wider (92.66 %) [11], and CSNN-blurr9 (92.85 %) [12]. *Niu et al.* [13], presents a thorough review summarizing and tracking the research progress of SNN for image recognition. Through this paper, we aim to establish a better methodology for image detection using snnTorch implemention.

## III. Proposed Methodology

snnTorch is a library that aims to bridge the gap between traditional deep learning approaches and biologically inspired neural computations. This allows researchers to explore applications in which conventional approaches already excel but SNN may be used to enhance efficiency and computational overhead such as image detection. snnTorch enables integration with PyTorch's autograd system which makes it seamless to incorporate SNNs into existing architectures. The architecture of snnTorch is built around core components that create a functional spiking neural network together. It commonly works on Leaky Integrate-and-Fire (LIF) neurons due to their simplicity and biological plausibility [1]. The working of snnTorch architecture starts with the input layer. This consists of neurons that receive data encoded as spikes. The encoding method allows network to process temporal information. The method works on frequency of spikes over time that represent intensity of values. Here, higher intensity corresponds to more frequent spikes. The output of this layer goes through to the hidden layers. It consists of multiple types of spiking neurons arranged in various configurations. Users can define these layers using PyTorch constructs in snnTorch while incorporating spiking behaviour through the neuron classes. The layer consisting of output neurons that generate spike trains corresponding to class predictions is the output layer. Fig. 1 and Fig. 2 represent the general architecture of SNN networks through snnTorch implementation [2].

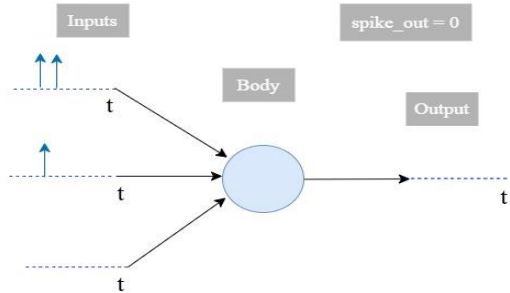

Fig. 1. Architecture of SNN using snnTorch with output = 0

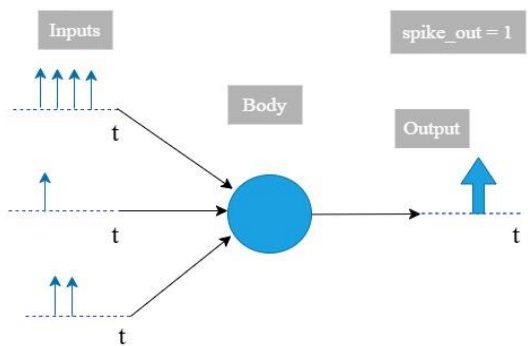

Fig. 2. Architecture of SNN using snnTorch with output = 1

These layers enable the training process. The training process generally goes through phases of data preparation, forward pass, loss calculation, backpropagation and optimization [14]. Data preparation corresponds to preprocessing and conversion of spike trains as input to the network. Forward pass follows this phase. The input spikes are fed through network layers during training wherein neuron's membrane potential is updates based on incoming spikes. The loss during the training is defined by a loss function which quantifies the difference between predicted outputs (spike patterns) and actual labels. This loss function is used to update synaptic weights using backpropagation governed by surrogate gradients in SNNs. To minimize the loss function iteratively, we perform optimization using optimizers. Fig. 3 represents the overall training process of SNN.

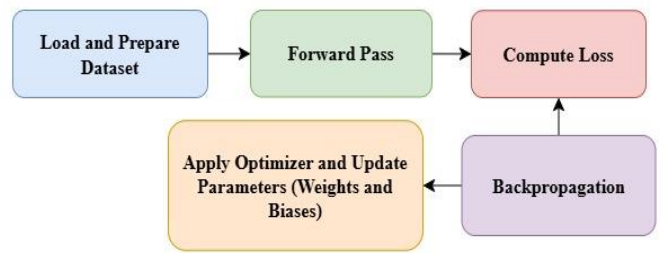

Fig. 3. The training process of SNN

Our model employs a similar training process for vehicle image detection that further emphasizes applications such real-time processing capabilities for autonomous vehicles, robotics and computer vision. These applications relied upon approaches such as CNNs traditionally [3]. The novelty of our project lies in integrating advanced techniques such as surrogate gradients and batch normalization within our model architecture that employs SNN using snnTorch implementation and leverage the unique temporal processing abilities to reduce computational overhead as well as achieve a performance on par with CNNs for image detection particularly vehicle image detection. We aim to utilize a multi-layer feedforward architecture for accomplish this objective. It would be composed of spiking neurons designed specifically for vehicle image detection tasks that make up three fully connected layers. These layers are applied with batch normalization after each linear transformation. Table I. represents the algorithm that we have employed for our task.

TABLE I. ALGORITHM FOR VEHICLE DETECTION USING SNN

| Algorithm 1 | Vehicle Detection using SNN |
|---|---|
| 1: | Initialize the dataset by loading images of vehicles and non-vehicles from directories. |
| 2: | Perform pre-processing (greyscale, normalization and resizing) on images. |
| 3: | Split the dataset into 80% training and 20% testing sets. |
| 4: | Create PyTorch DataLoader object to handle batch processing. |

5: Define key hyperparameters and initialize SNN model with required specifications for input layer, LIF neurons with fast_sigmoid surrogate gradient, two hidden layers and output layer with the hyperparameters.
6: Define the loss function as cross-entropy rate loss and initialize AdamW optimizer with learning rate of 1e-3.
7: Set early stopping parameters with patience set to 5 epochs.
8: **For** each epoch from 1 to 20 **do**:
9:     Train SNN on training dataset:
10:         Reset gradients.
11:         Perform forward propagation for 50 time steps.
12:         Compute cross-entropy loss using spike recordings.
13:         Perform backward propagation to compute gradients.
14:         Update model parameters.
15:         Record training loss and accuracy.
16:     Evaluate SNN on testing dataset:
17:         Compute loss and accuracy.
18: **End for**
19: Update best_loss according to current test loss.
20: Save the best model and load for evaluation.
21: Calculate final evaluation metrics and visualize metrics.

It consists of:

- Input layer: Normalizes input data from images encoded as spike trains.
- Two Hidden Layers: Utilize Leaky Integrate-and-Fire (LIF) neurons with surrogate gradients.
- Output Layer: Produces spike outputs corresponding to predicted classes (vehicle or non-vehicle).

The training process employs a learning rate of 1e-3. The learning rate controls the step size of the parameter updates during gradient descent. The parameter update rule for a parameter $\theta$ using learning rate $\eta$ is:

$$\theta_{t+1} = \theta_t - \eta \nabla L(\theta_t) \qquad (1)$$

where $\nabla L(\theta_t)$ is the gradient of the loss function with respect to $\theta$ at time step $t$ [15].

The learning rate is used with the AdamW optimizer for stable convergence. It is a variant of Adam optimizer with decoupled weight decay, improving generalization by preventing large weights. The update rule for AdamW is:

$$m_t = \beta_1 m_{t-1} + (1 - \beta_1)\nabla L(\theta_t) \qquad (2)$$

$$v_t = \beta_2 v_{t-1} + (1 - \beta_2)(\nabla L(\theta_t))^2 \qquad (3)$$

$$\hat{m} = \frac{m_t}{1 - \beta_1^t} \qquad (4)$$

$$\hat{v} = \frac{v_t}{1 - \beta_2^t} \qquad (5)$$

$$\theta_{t+1} = \theta_t - \eta \frac{\hat{m}_t}{\sqrt{\hat{v}_t} + \epsilon} - \lambda \theta_t \qquad (6)$$

where $\beta_1$ and $\beta_2$ are exponential decay rates, $\lambda$ is the weight decay factor and $\epsilon$ is a small constant for numerical stability [16].

We couple this learning rate and optimizer with the fast_sigmoid activation function of snnTorch.

Spiking neurons output discrete spike (non-differentiable functions); thus, backpropagation cannot be directly applied and fast_sigmoid, a type of surrogate gradient function is used during training to approximate the gradient of the spike function. The fast_sigmoid surrogate gradient function is defined as:

$$\sigma'(u) = \left(\frac{1}{1 + |u|}\right)^2 \qquad (7)$$

where $u$ is the membrane potential of the neuron [17].

To control how quickly the membrane potential of a neuron decays over time in the Leaky Integrate-and-Fire (LIF) model, we use decay factor. The values correspond to the retention of information. A higher value means the decay is slower, which allows the neuron to retain information longer. We have defined decay factor as BETA in our model with the value of 0.95. The update of membrane potential $V$ of a neuron at time step t is updated as:

$$V_t = \beta V_{t-1} + I_t \qquad (8)$$

where $\beta$ is the decay factor, and $I_t$ is the input current at time step $t$ [15].

Further, we have defined the loss function as Cross-Entropy Rate Loss as it is designed specifically for spiking neural networks where the outputs are spike trains rather than continuous values. This loss function computes the cross-entropy between the predicted spike rate and the true labels over multiple time steps. Given the predicted spike rate $\hat{y}$ and true label $y$, the cross-entropy loss is:

$$L = -\sum_{i=1}^{C} y_i \log(\hat{y}_i) \qquad (9)$$

where $C$ is the number of classes [2].

Additionally, an important factor helped us enhance and influence the network toward the correct direction during the training process. We used Batch Normalization BatchNorm1d to normalize the input of each layer to stabilize learning during training process and accelerate convergence. This is proved to be useful for spiking networks where activations vary significantly over time

steps. For a batch input $x$ with mean $\mu$ and variance $\sigma^2$, batch normalization is defined as:

$$\hat{x}_i = \frac{x_i - \mu}{\sqrt{\sigma^2 + \epsilon}} \quad (10)$$

The normalized output is then scaled and shifted:

$$y_i = \gamma \hat{x}_i + \beta \quad (11)$$

where $\gamma$ and $\beta$ are learnable parameters [18].

Finally, we evaluate the performance of SNN trained through various evaluation metrics such as accuracy and confusion matrix. Thus, through innovative choices and hyperparameter setting aimed at enhancing performance and energy efficiency, we provide a comprehensive framework through which snnTorch may be utilized to construct SNN for vehicle image detection.

## IV. RESULTS AND DISCUSSION

### A. Dataset

The dataset we used in this research is the "Vehicle Detection Image Set", extracted from Kaggle [19]. The dataset has two classes, Vehicle and Non-Vehicle, that add up to 17,760 images; out of which we utilized around 1070 images in both the classes. The images were pre-processed to maintain uniformity by resizing them to 128x128 pixels while at the same time preserving their aspect ratio. Moreover, padding was applied to maintain a consistent square shape for all images, with grayscale conversion to simplify the processing. The pixel values were normalized to the range [0, 1] for enhancing model compatibility. In order to get a balanced distribution during training, the pre-processed dataset was randomly shuffled.

For evaluating and training our mode, we split the data into training and testing sets, assigning 80% and 20% to the sets respectively. The splitting helps in substantiating the model's generalizability and predictive powers.

### B. Hyperparameters and Results

The hyperparameters play a very important role in influencing the model's efficacy and overall performance. As shown in Table II., we use a batch size of 30 that helps in balancing the computational efficiency and stability of the gradient. A learning rate of 1e-3 fits best for our model ensuring steady convergence during training, with the 20 epochs to allow sufficient iterations for the model to learn. A hidden layer of size 64 lets the model adequately learn the complex patterns, while the membrane potential decay rate of 0.95 maintains a balance between spiking activity and temporal memory. The number of simulation steps per input is set to 50 are carefully tuned to optimize the SNN model.

The categorical cross-entropy rate loss function is used in the training process which leads the model to minimize errors by calculating the difference between predicted and actual outputs. The AdamW optimizer which is known for its ability to balance weight updates along with weight decay, is also used for ensuring efficient convergence while at the same time preventing overfitting.

TABLE II. HYPERPARAMETERS AND THEIR VALUES USED

| Hyperparameters | Values |
|---|---|
| Batch Size | 30 |
| Learning Rate | 1e-3 |
| Number of Epochs | 20 |
| Hidden Layer Size | 64 |
| Beta | 0.95 |
| Number of Steps | 50 |
| Input Image Size | 128x128 |
| Patience | 5 |

The training of the SNN model also utilizes early stopping based on the test loss, with patience set to 5, which saves the best model at epoch 9. Here, it achieved the lowest test loss of 0.1665 and a test accuracy of 94.82%. This early stopping prevents overfitting by stopping the training after epoch 14. There is a consistent improvement in both training and test accuracies throughout the initial epochs, which indicates that the model is generalizing well on the test set. In fig. 4, the plot showing loss versus epochs projects a downward trend in training loss, which indicates that the model is getting better at minimizing errors as it learns. The test loss fluctuated a bit after the initial epochs but eventually stabilized, confirming that early stopping was effective in retaining the best model. At the same time, fig. 5 presents the accuracy versus epochs graph, which shows a uniform growth in both training and test accuracies. A significant increase in test accuracy at epoch 9 along with the lowest test loss, shows that the model performs best at this point. Following this, evaluation using the Receiver Operating Characteristic (ROC) curve as corroborated in fig. 6, validates the model's strong distinguishing power, achieving an Area Under the Curve (AUC) score of 0.99. The ROC curve hits close to the top-left corner underscores excellent sensitivity and specificity. Additionally, the confusion matrix provides an exhaustive analysis of classification results in fig. 7. Out of 425 test samples, the model correctly classifies 216 Non-Vehicle images and 195 Vehicle images, with only 9 false positives and 5 false negatives, thus highlighting the robustness of the trained SNN model.

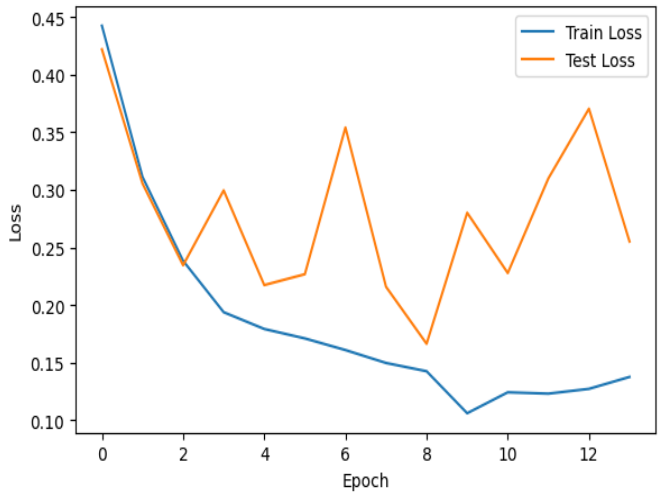

Fig. 4. Graph of Loss vs. Epochs

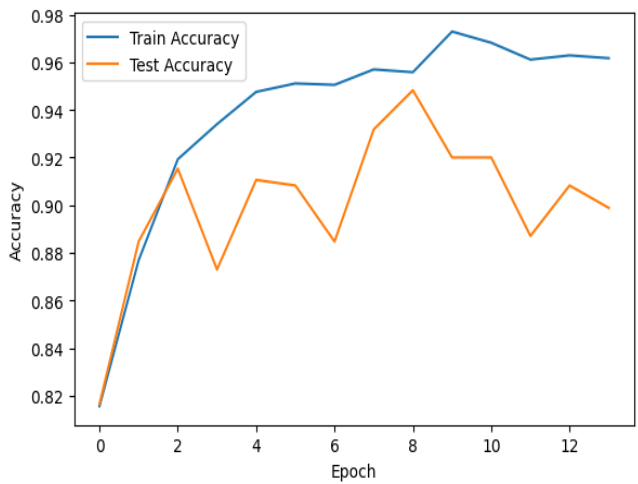

Fig. 5. Graph of Accuracy vs. Epochs

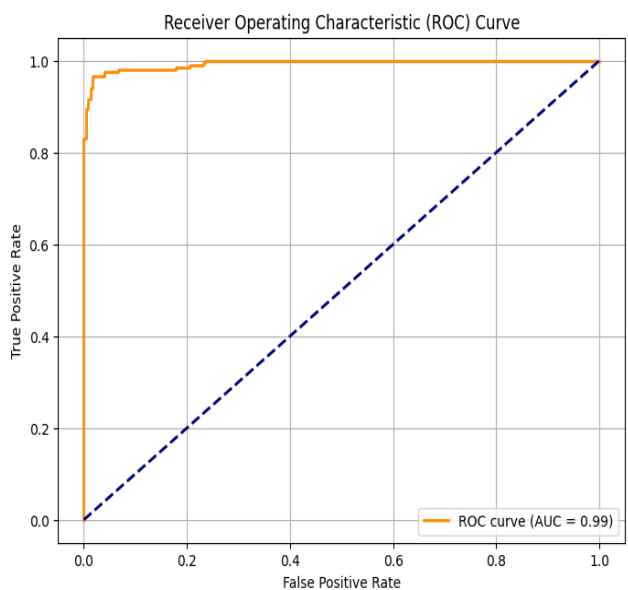

Fig. 6. ROC Curve

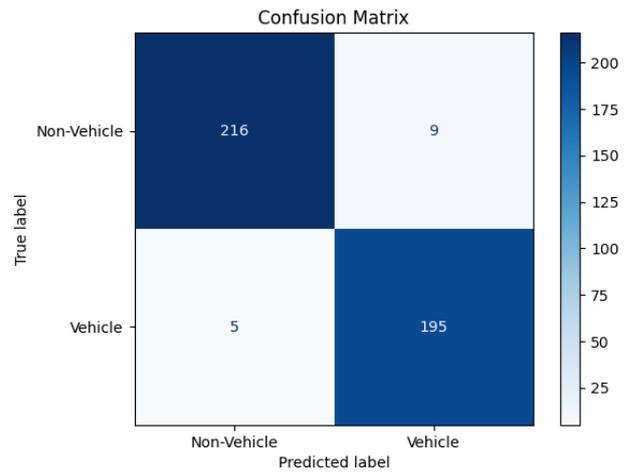

Fig. 7. Confusion Matrix

The performance metrics shown in Table III, provide a qualitative comparison of DRiVE with existing models used for image detection. Our model's accuracy outperforms some of the state-of-the-art SNN models, underscoring its robustness and precision in the task of detection.

TABLE III. COMPARISON OF PERFORMANCE METRICS WITH OTHER SNN MODELS

| SNN Models | Accuracy |
| --- | --- |
| DRiVE | 94.82% |
| AMOS [9] | 80.97% |
| CSNN-blurr9 [12] | 92.85% |
| SpikeformerV2 [10] | 94.80% |
| S-ResNet38 wider [11] | 92.66% |

### C. Discussion

Based on our results, implementing our model, DRiVE, exhibits a reliable and novel approach for vehicle detection utilizing the snnTorch mechanism. The size of the hidden layer in our model plays a very crucial part in achieving high accuracy. Added to this, it directly affects the model's ability to grasp intricate patterns. The excellent AUC score DRiVE achieves also underscores its power to learn to differentiate between vehicles and non-vehicles, rather than just hit a wild guess. The promising results of our study open the way for industrial applications like autonomous vehicles, traffic monitoring, highway surveillance, and toll collection systems.

## V. CONCLUSION

To summarize, our study showed the breakthrough potential of the snnTorch mechanism in transforming image detection within the Spiking Neural Networks. Our model DRiVE has consistently demonstrated excellent predictive accuracy and ability to differentiate between two classes, asserting its readiness in industrial scenarios. Moreover, an accuracy of 94.82% and an AUC score of 0.99 further underpin its robustness and its discerning ability. It paves the way for future research to explore more use cases using the snnTorch model. With its proven performance, DRiVE establishes itself as a reliable framework for advancing real-world applications in fields like autonomous systems and intelligent transportation.